%% file: iclr2024_conference.tex
\documentclass{article} 
\usepackage{iclr2024_conference,times}
\usepackage{graphicx}
\usepackage{comment}
\usepackage{caption}
\usepackage{subcaption}
\input{math_commands.tex}

\usepackage{hyperref}
\hypersetup{
    colorlinks=true,
    linkcolor=blue,
    citecolor = blue,
    filecolor=magenta,      
    urlcolor=cyan,
    pdftitle={},
    pdfpagemode=FullScreen,
}
\usepackage{url}

\title{Causal Graph Neural Networks for Wildfire Danger Prediction}

\author{Shan Zhao$^{1}$, Ioannis Prapas$^{2,3}$, Ilektra Karasante$^{2}$, Zhitong Xiong$^{1}$, Ioannis Papoutsis$^{2}$, \\
\textbf{Gustau Camps-Valls$^{3}$, Xiao Xiang Zhu$^{1}$} \\
$^{1}$ Technical University of Munich, Ottobrunn, 85521, Germany \\
$^{2}$ National Observatory of Athens, Athens, 11810, Greece \\
$^{3}$ University of València, Valencia, 46980, Spain }

\newcommand{\ZS}[1]{\textcolor{black}{#1}}

\iclrfinalcopy 
\begin{document}

\maketitle

\begin{abstract}
Wildfire forecasting is notoriously hard due to the complex interplay of different factors such as weather conditions, vegetation types and human activities. Deep learning models show promise in dealing with this complexity by learning directly from data. However, to inform critical decision making, we argue that we need models that are right for the right reasons; that is, the implicit rules learned should be grounded by the underlying processes driving wildfires. In that direction, we propose
integrating causality with Graph Neural Networks (GNNs) that explicitly model the causal mechanism among complex variables via graph learning. The causal adjacency matrix considers the synergistic effect among variables and removes the spurious links from highly correlated impacts. Our methodology's effectiveness is demonstrated through superior performance forecasting wildfire patterns in the European boreal and mediterranean biome. The gain is especially prominent in a highly imbalanced dataset, showcasing an enhanced \ZS{robustness} of the model to adapt to regime shifts in functional relationships. Furthermore, SHAP values from our trained model further enhance our understanding of the model's inner workings. 

\end{abstract}

\section{Introduction}
Wildfires present considerable risks to human safety and economic stability in affected regions and wildlife habitats. \ZS{The risk is inflated by climate change, which aggravates wildfires, particularly in the vulnerable european ecosystems that this study targets \citep{batllori_climate_2013}. 
Accurately predicting the highly favorable conditions for a wildfire} is vital for effective disaster prevention and preparedness. Earth observation data is pivotal in fire danger assessment and forecasting, offering extensive meteorological and vegetation condition information closely related to wildfire occurrences \citep{pettinari2020fire}. The copious amount of data available lends itself well to sophisticated deep learning (DL) approaches \citep{CampsValls21wiley,reichstein2019deep} for predicting wildfire danger \citep{prapas2021deep}. Recent advancements have been made by taking multiple factors as model inputs. \citet{li2023attentionfire_v1} underscore the importance of delayed interactions between wildfires and climate patterns in the forecasting scenarios. \citet{kondylatos2022wildfire} identified the most critical predictors of the next day's wildfire danger using explainable AI (xAI) methods. TeleVit \citep{prapas2023televit} incorporates different local, global variables with teleconnection indices within a transformer-based architecture.  However, these approaches do not explicitly model the information flow within the model. Consequently, the mechanism of how the variables interact and assist the final prediction is not considered during the model design and training procedure. 
\par
\ZS{The application of causal inference in Remote Sensing enhances our understanding of processes in the Earth system and facilitates effective interventions \citep{runge2023causal,camps2023discovering}. \citet{perez2018causal} propose an asymmetry-based approach that uses sensitivity criterion to address the causal direction between geoscientific variables. State-space method CCM \citep{tsonis2018convergent} finds its applications in Ecosystem \citep{sugihara2012detecting}, land-atmosphere interactions \citep{wang2018detecting,diaz2022inferring}, and identifying influencing factors in urban soil pollutants \citep{gao2023causal}.} Integrating causal inference with DL architectures remains challenging due to the gap between the low-level inputs (e.g. images or earth observation data) and high-level definitions of graphic nodes. \ZS{ANM \citep{zhang2012identifiability} uses Multi-layer Perceptions to estimate the nonlinear causal effects between variables. \citet{iglesias2023causally} propose a Causal Neural Network (NN) to improve climate projection. However, most efforts are limited to feature selection and observational data regression.}
\par
Intricate causal relationship modeling among variables can support robust and reliable DL \ZS{\citep{scholkopf2021toward,varando2021learning}}. In this work, we incorporate a causality graph into the GNN architecture, thereby capturing the information flow through graph learning techniques. Our experimental results reveal that the causality-infused graph is a robust indicator of pertinent information, enhancing prediction quality. The SHapley Additive exPlanations (SHAP) \citep{lundberg2017unified} values derived from our model reveal the memory effect of teleconnections in driving the wildfire danger.

\section{Methods}
\begin{figure}[tb]
    \centering
    \includegraphics[width=0.8\columnwidth,  clip, trim={1.7cm 2.9cm 2.4cm 1.9cm}]{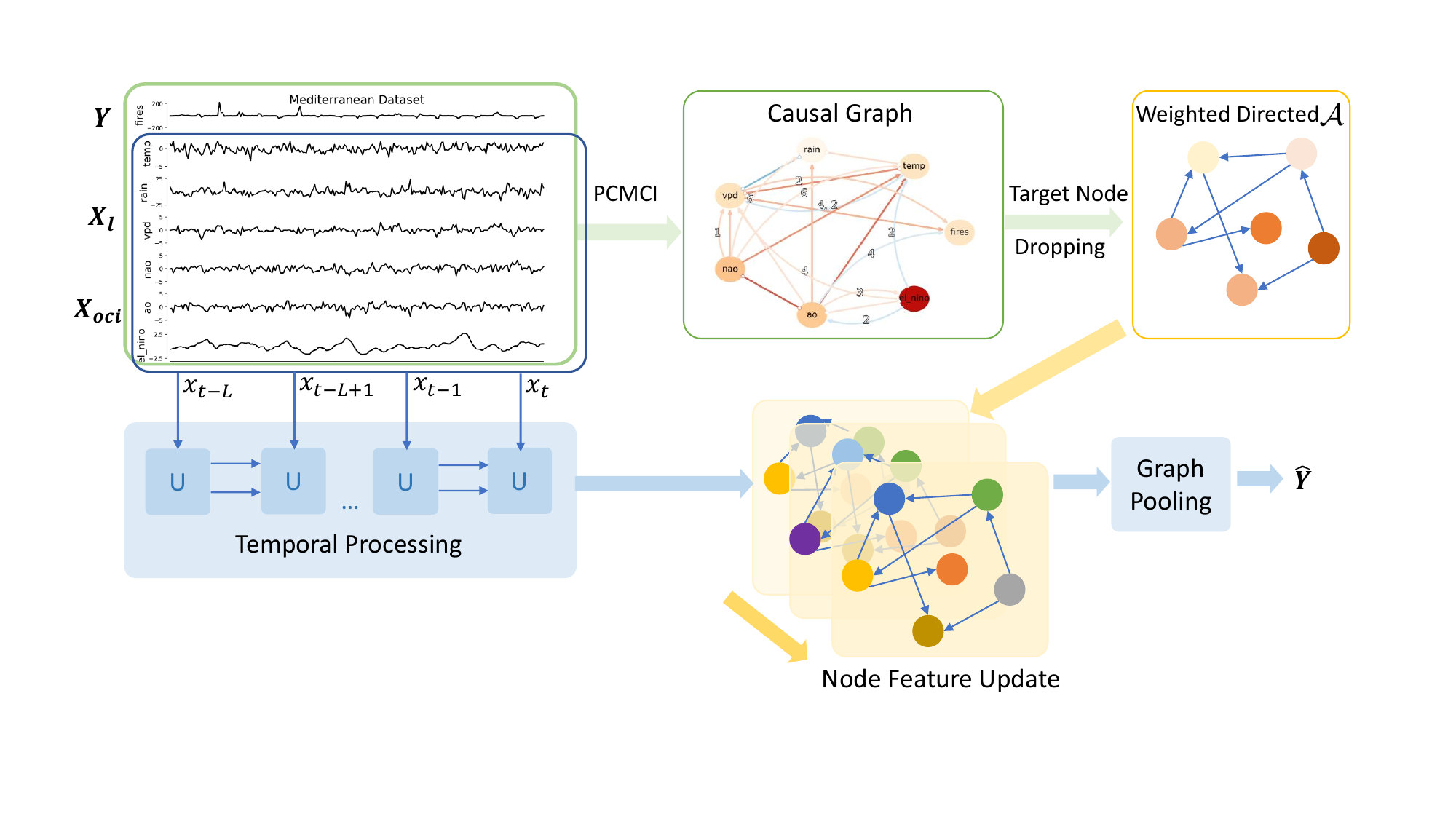}
    \caption{\small Workflow of our proposed Causal-GNN. The inputs contain local and OCI variables with various temporal and spatial scales. The adjacency matrix captures the causal relationship among variables. The node feature is extracted by the temporal module and updated via GNNs for the final prediction. The cross-entropy between the prediction and ground truth is minimized.}
    \label{fig:workflow}
\end{figure}
In the Earth system, \textit{Synergistic Effects} \ZS{\citep{runge2019inferring}} are the combined effects of drivers making an impact that is greater than the simple sum of its parts. For instance, devastating wildfires are often related to dryness conditions, combustible materials, and an ignition source, but \ZS{their} impacts are subjected to threshold behavior \citep{reichstein2013climate}. Causal methods allow us to identify regime shifts in functional relationships triggered by extreme conditions \citep{runge2019inferring}. We consider leveraging the causal graph and the message passing of a temporal GNN to capture the wildfire dynamics.
\par
Given \ZS{an input of} local scale variables $X_{l} \in \mathbb{R}^{B\times C_l \times L_l}$ and Oceanic and Climatic Indices (OCIs) variables $X_{oci} \in \mathbb{R}^{B\times C_{oci} \times L_{oci}}$, where $B$ is the batch size, $C$ is the number of variables, and $L$ is the time lag to the current time step $t$\ZS{, our} task is to predict the wildfire danger $Y \subset \{0,1\} \in \mathbb{R}^{B}$ in the future time step $t+h$, where $h$ is the forecasting horizon.
\par
GNNs extend DL to data with a graph structure. The two key components of GNNs are the adjacency matrix and node features. Given that causality is usually represented with graphs \citep{runge2019detecting}, we create this adjacency matrix $\mathcal{A} \in \mathcal{R}^{(C_l+C_{oci}) \times (C_l+C_{oci})}$ from a causal graph. The strength of the causal link quantifies weights of $\mathcal{A}$. The causal graph is the graphic representation of the causal relationship among the variables. Our selection algorithm to compute the causal graph is the PCMCI method \citep{runge2019detecting}. The nodes in the time series causal graph represent the variables $X_t^j$ at different lag times, and $\mathcal{P}a(X_t^j) \in \mathbf{X}_t^- = \mathbf{X}_{t-1}, \mathbf{X}_{t-2}, ...$ denotes the causal parents of variable $X_t^j$. A causal link $X_{t-\tau}^i \rightarrow X_t^j$ exits if $X_{t-\tau}^i \in \mathcal{P}a(X_t^j)$. Multiple factors, such as a temporal trend, geographical shifts, and vegetation type, can introduce non-stationary dependencies that act as confounders. Thus, the data is preprocessed to ensure the causal stationarity. The preprocessing is detailed in Appendix \ref{subsec:causal_stationarity}. However, due to finite sample length and partially unfulfilled standard assumptions of causal discovery \citep{Pearl2009,spirtes2001causation}, spurious links can still exist \citep{Krich2020}. To validate the findings with domain knowledge, we employ link assumptions (Appendix \ref{subsec:link_assumption}) to orient parts of the edges.
\par  
Regarding the vertices in GNN, each node represents the temporal feature of the given variable, which is extracted by a single layer LSTM \citep{memory2010long}. The node feature is updated by message transmission among neighboring nodes whose connectivity is defined within the adjacency matrix. The node feature is refined via multiple convolution layers with layer norm and activation. The graph pooling at the last layer of GNN gathers the global information from the whole graph. The details of the model are in Appendix \ref{subsec:model_details_hyperparameters}. The architecture is shown in Fig.\ref{fig:workflow}.
\section{Experiments}
The dataset used is the SeasFire Datacube \ZS{\citep{karasante2023seasfire}}, a global dataset spanning 21 years, featuring an 8-day temporal and $0.25 \deg$ spatial resolution, containing crucial information on seasonal wildfire drivers, targets, and masks. We select three local weather variables Temperature at 2 meters - Mean (T2m), Total precipitation (TP), Vapor Pressure Deficit (VPD), and three Oceanic Climate Indices (OCIs) known to influence the test area, including North Atlantic Oscillation (NAO), Arctic Oscillation Index (AO) \citep{MllerPlath2022}, and Nino 3.4 Anomaly (Nino34\_anom) \citep{Brnnimann2007}. The time lag is 10 months for OCIs and 39 $\times$ 8 days for local variables. The target variable is a binary map of the burned areas. Additionally, we narrow the spatial coverage to Europe and test two cases with biome types ``Mediterranean'' and ``Boreal'' according to the static features of GFED regions and biomes. The causal graph contains seven nodes with the target variable included. The $\mathcal{A}$ masks out edges and nodes related to the target variable to avoid label leakage. The NNs use 2003 to 2015 data for training, 2016 to 2017 for validation, and 2018 to 2019 for testing. We compare our proposed model to the following baselines: 1) Long short-term memory (LSTM) \citep{memory2010long}. 2) Gated recurrent units (GRUs) \citep{cho2014learning}. 3) GNN\_CORR uses the correlation coefficient (CC)-based adjacency matrix computed from temporal features of variables. 4) GNN\_FULL uses a fully connected graph as the adjacency matrix. The model’s performance is evaluated using the Area Under the Precision-Recall Curve (AUPRC) and the Area Under the Receiver Operating Characteristic Curve (AUROC). We test \ZS{models' forecasting performance eight days in advance in the Mediterranean, and 1, 2, 8 $\times$ eight days in Boreal to explore their long-term capacity}.
\section{Results and Analysis}
\begin{figure}[htb] 
\centering
\begin{subfigure}{.5\textwidth}
  \centering
  \includegraphics[width=0.9\columnwidth]{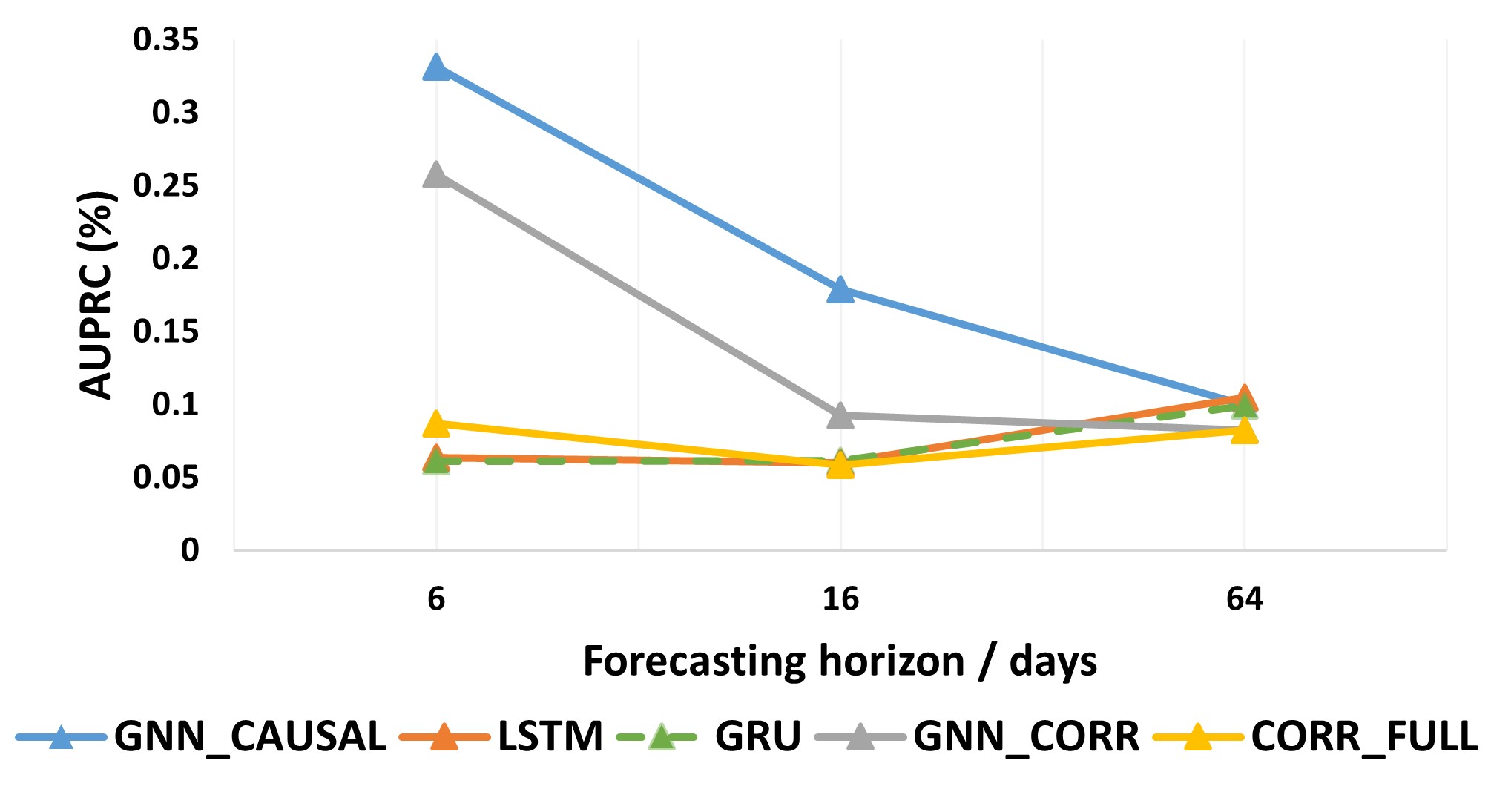}
  \caption{}
    \label{fig:biomes9}
\end{subfigure}%
\begin{subfigure}{.5\textwidth}
  \centering
  \includegraphics[width=0.75\linewidth]{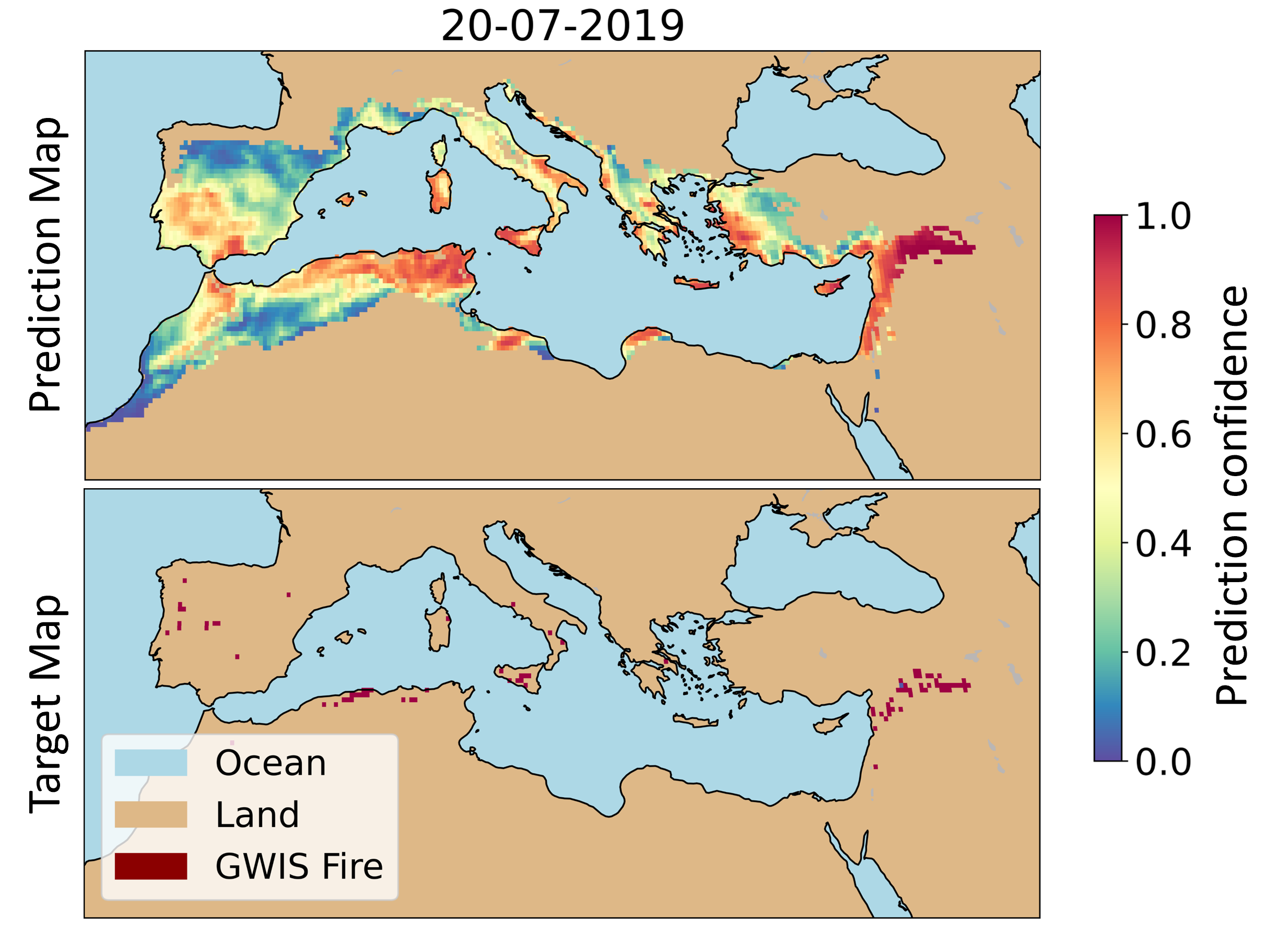}
  \caption{}
  \label{fig:fireDangerMap}
\end{subfigure}
\caption{\small (a) AUPRC performance of the different models at forecasting horizon of 8, 16, 64 days in Boreal. (b) Causal-GNN predicted the sample fire danger map at eight days lead forecasting time in the Mediterranean.}
\label{fig:test}
\end{figure}

The model performance at the forecasting horizon of 8 days in the Mediterranean is listed in Table \ref{tab:biomass3}. The baseline AUPRC achieved by a Random classifier equals the fraction of positives \citep{saito2015precision} in the dataset, which is around 1.1\%. 
The causal GNN shows competitive performance with common time-series classification methods and improved performance over a fully connected and CC-based graph. This highlights the importance of removing spurious links during GNN training, where the redundant edges can lead to over-smoothed node features and degraded prediction accuracy. The prediction map Fig.\ref{fig:fireDangerMap}, essentially looks like a fire danger map as in similar studies which approach fire danger forecasting as a classification task. It matches well the target burned areas in a given 8-day period of summer 2019, which is part of the test set.
\begin{table}[htb]
\small
    \centering
    \begin{tabular}{l c c c}
        \hline
        Model & AUPRC (\%) $\uparrow$& AUROC (\%) $\uparrow$ \\ \hline
         Random classifier & 1.1 & 50.0 \\
         LSTM & 32.9 & 91.9\\
         GRU & \textbf{34.6} & 91.6 \\
         \hline
         GNN\_CORR & 30.5 & 92.1\\
         GNN\_FULL & 29.1 & 90.8\\
         GNN\_CAUSAL (Proposed) & \textbf{34.6} & \textbf{92.4}\\ \hline
    \end{tabular}
    \caption{\small AUPRC and AUROC performance of the different models for $h$ = 8 days in Mediterranean.}
    \label{tab:biomass3}
\end{table}
The model performance at different prediction horizons in Boreal is shown in Fig.\ref{fig:biomes9}. The proportion of positive samples in this area is 0.0737\%, a much more challenging case for all DL models. Our \ZS{m}odel achieves the best AUPRC at the short horizon. However, with the increasing horizons, the performance of Causal-GNN deteriorates with the rest of the models. \ZS{The model's limited long-term forecasting capacity may be attributed to its static causal graph. Incorporating time-series graphs to account for time lags could potentially mitigate this issue.}

\par
\begin{figure}[htb] 
\centering
\begin{subfigure}{.5\textwidth}
  \centering
  \includegraphics[width=0.8\linewidth, clip, trim={0cm 0.5cm 0cm 0.5cm}]{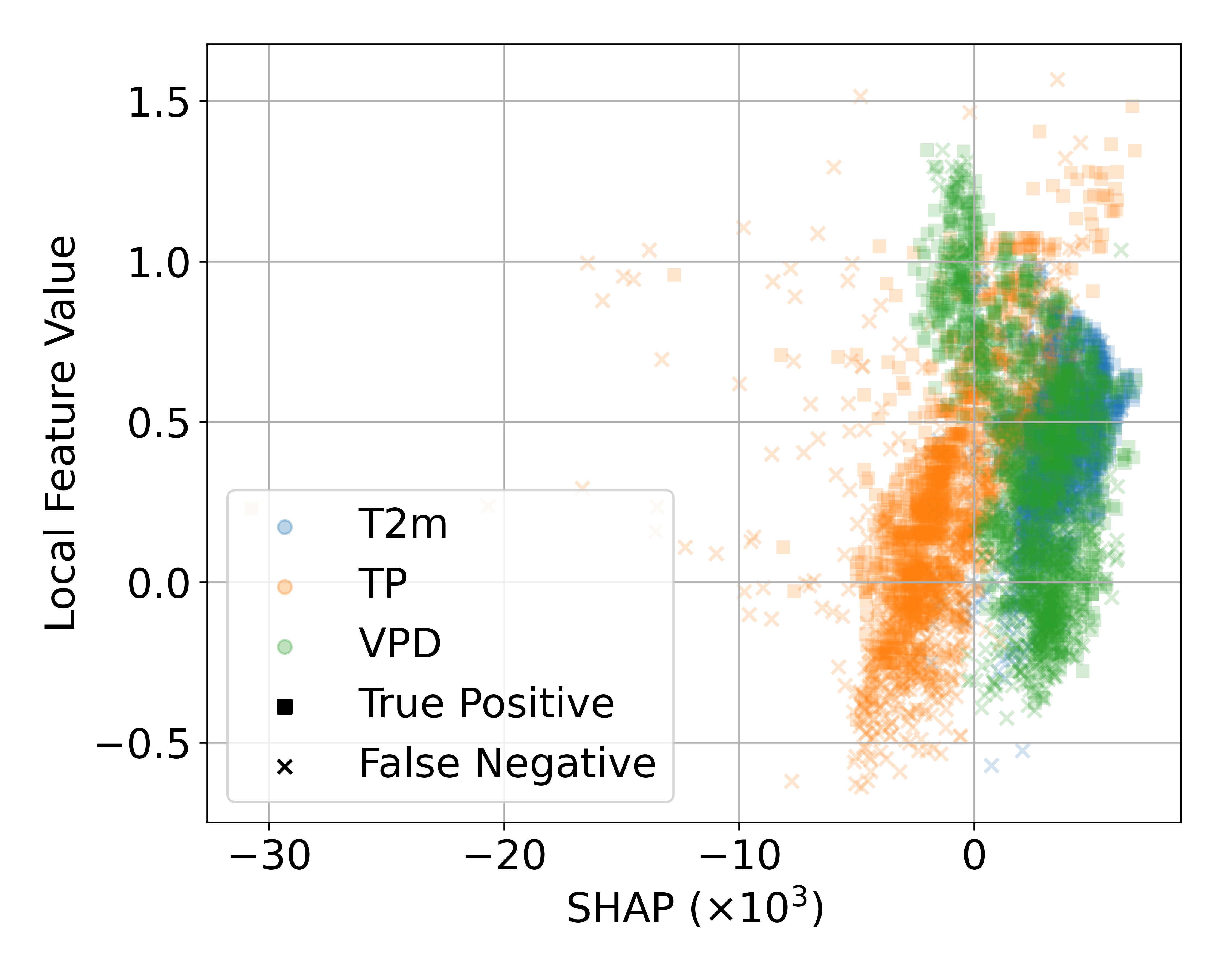}
  \caption{\small SHAP value w.r.t local feature values.}
  \label{fig:shap_local}
\end{subfigure}%
\raisebox{0.1\height}{
\begin{subfigure}{.5\textwidth}
  \centering
  \includegraphics[width=\linewidth, clip, trim={0cm 2.5cm 1cm 3.6cm}]{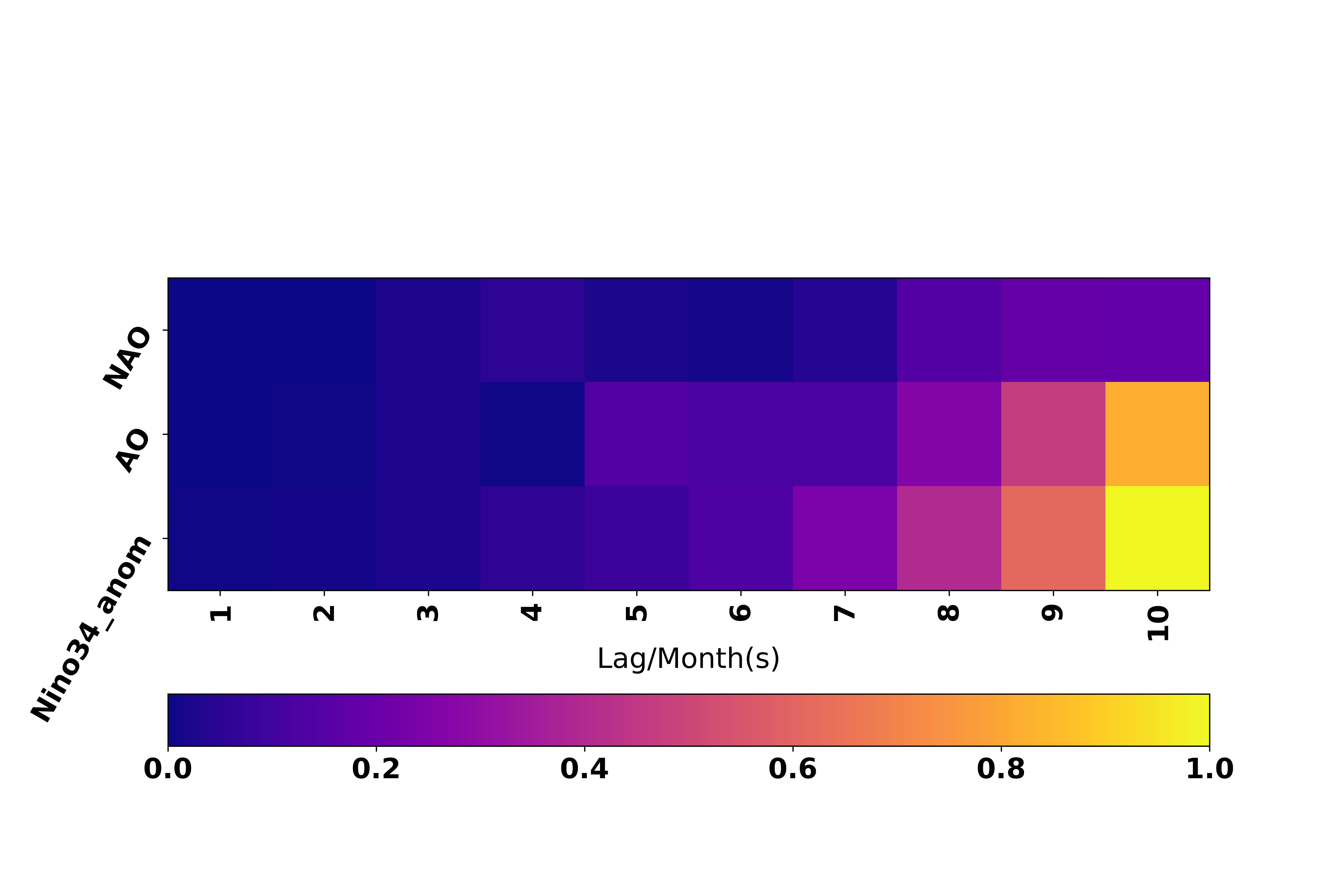}
  \caption{\small Scaled absolute SHAP value of OCIs with changing lags.}
  \label{fig:shap_oci}
\end{subfigure}}
\caption{\small (a) Square (cross) markers are True Positive (False Negative) samples. Positive (negative) SHAP value contributes to higher (lower) fire danger. (b) The warmer color shows a higher impact on the prediction.}
\label{fig:test}
\end{figure}

\ZS{Understanding the choice of Causal-GNN helps to refine the model and verify its physical consistency. Thus, w}e conduct SHAP analysis on positive test samples in the Mediterranean. From Fig.\ref{fig:shap_local}, we notice that high temperatures increase wildfire danger. In areas where the VPD is low ($<$0.7), the drier conditions correlate with an increased likelihood of wildfire. This relationship inverts while at the high VPD ($>$0.7) part. 
Figure~\ref{fig:shap_oci} is the absolute SHAP value of OCIs at various lags averaged over samples. The increased value on large lags signifies the memory effects of large-scale atmospheric circulation patterns in driving future wildfires. For example, El Ni$\Tilde{\text{n}}$o intensifies the wildfire danger, especially six to ten months in advance.
\section{Conclusion}
This work proposes a DL framework that integrates the causal structure into the GNN training. The causal structure reveals the intricate interplay among variables and the temporal GNN, which takes the temporal feature of variables as nodes and weighted causal graph as connectivity, promoting the information exchange between variables that simulate both regional and larger-scale earth system dynamics. The evaluation of two cases with different vulnerable forest types shows promising results, especially when the dataset is highly imbalanced, indicating the robustness brought by causal knowledge. The gain over the CC-based method shows its efficacy in considering the synergistic effect and spurious link removal. xAI reveals the physically-consistent dependencies in the learned model. \ZS{Our method paves the way to} enhance process understanding and \ZS{model trustworthiness}.  \ZS{Future efforts will focus on developing learnable causal graphs and examining the global variability. Additionally, we plan to explore regression of the burned area, with the goal to improve our understanding of different wildfire types and especially extreme events.}
\section{Acknowledgments}
The work of I. Prapas, I. Karasante and I. Papoutsis is funded by the European Space Agency in the context of the SeasFire project. The work of S. Zhao, Z. Xiong, and XX. Zhu is jointly supported by the German Federal Ministry of Education and Research (BMBF) in the framework of the international future AI lab "AI4EO -- Artificial Intelligence for Earth Observation: Reasoning, Uncertainties, Ethics and Beyond" (grant number: 01DD20001), by German Federal Ministry for Economic Affairs and Climate Action in the framework of the "national center of excellence ML4Earth" (grant number: 50EE2201C), by the Excellence Strategy of the Federal Government and the Länder through the TUM Innovation Network EarthCare and by Munich Center for Machine Learning.

\bibliography{iclr2024_conference}
\bibliographystyle{iclr2024_conference}

\newpage
\appendix
\section{Model Details and Hyperparameters}\label{subsec:model_details_hyperparameters}
The samples in the training and validation set \ZS{are resampled} so that the ratio of the number of negative samples (No fire) to positive samples (Fire) is 5 to 1. All learning rates are set as 1e-5 with a weight decay of 5e-6. Model weights are initialized using Xavier normalization. In the Causal-GNN, the temporal features of each input variable (channel-wise) are extracted by a single LSTM layer with the hidden state of dimension 256. The adjacency matrix is normalized to stabilize the graph learning. The temporal node feature is updated by a two-layer convolutional neural network with kernels \ZS{$K \in \mathbb{R}^{256 \times 512 \times 1 \times 1}$ and $\mathbb{R}^{512 \times 256 \times 1 \times1}$}, following LayerNorm and LeakyReLU. The Graph Pooling averages node features, leading to the final classification layer. This is a single linear layer with dimensions (256 and 2) for binary classification tasks. Confidence is determined by the softmax function of the positive prediction, and it is used to visualize the fire danger map in Fig.\ref{fig:fireDangerMap}.
\par
For the baselines, LSTM and GRUs project the input feature to a dimension of 256. We utilize an adjacency matrix $\mathcal{A}$ for the fully connected graph, wherein all elements are set to 1, indicating complete connectivity. The correlation coefficient-based graph employs PyTorch's \textit{corrcoef} function to calculate the similarity among variables within the temporal feature space. All performance scores are reported as the mean of three runs with different seeds.
\section{Causal Discovery}
\subsection{Causal Stationarity}\label{subsec:causal_stationarity}
\begin{figure}[htb] 
    \centering
    \includegraphics[width=0.6\textwidth]{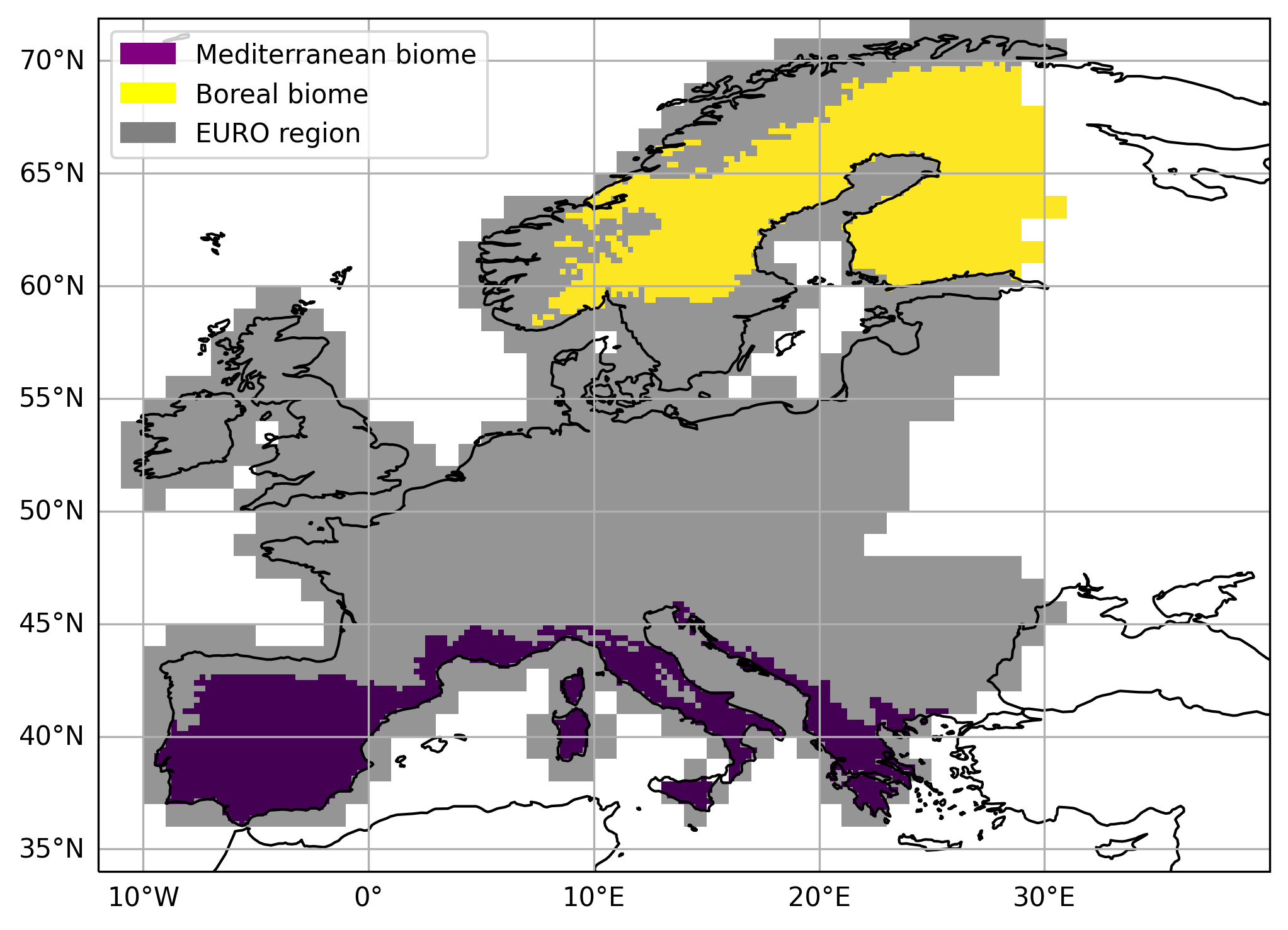} 
        \caption{Test biomes in EURO region.}
        \label{fig:biomes}
\end{figure}

To prepare for the spatial-temporal stationary dataset, the input local variables are resampled to one-month intervals, and their mean is removed from the inputs. We test two cases over the Europe area, which are located between 25$^{\circ}$N - 75$^{\circ}$N, 13$^{\circ}$W - 45$^{\circ}$E. We use the biomes 3.0 area and biomes 9.0 area of the gfed region 6.0 (EURO) to consider the fuel types of Europe's Mediterranean and Boreal biomes. Their geographical extension is shown in Fig.\ref{fig:biomes}.

\subsection{Link Assumption}\label{subsec:link_assumption}
For the time series, we assume time order, the causal Markov condition, faithfulness, causal sufficiency, causal stationarity, and no contemporary causal effects \citep{runge2019detecting}.

\begin{figure}[htb] 
    \centering
\includegraphics[width=0.5\textwidth]{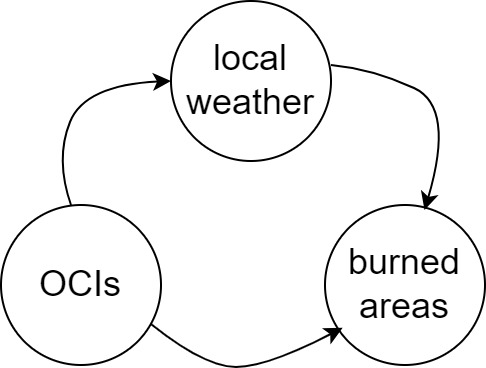} 
        \caption{Causal order in PCMCI ParCorr independence test. The local weather includes temperature, total precipitation, and vapor pressure variables. The OCIs are the Arctic Oscillation, the North Atlantic Oscillation, and El-Ni\~no in the 3.4 region. Here, the local weather acts as a mediator variable, explaining the relationship between an independent variable (OCIs) and a dependent variable (burned areas).}
        \label{fig:mediator}
\end{figure}

The link order is depicted in Fig.\ref{fig:mediator}.

\subsection{Causal Graphs}\label{subsec:causal_graphs}
\begin{figure}[htb] 
    \centering
    \includegraphics[width=1\textwidth]{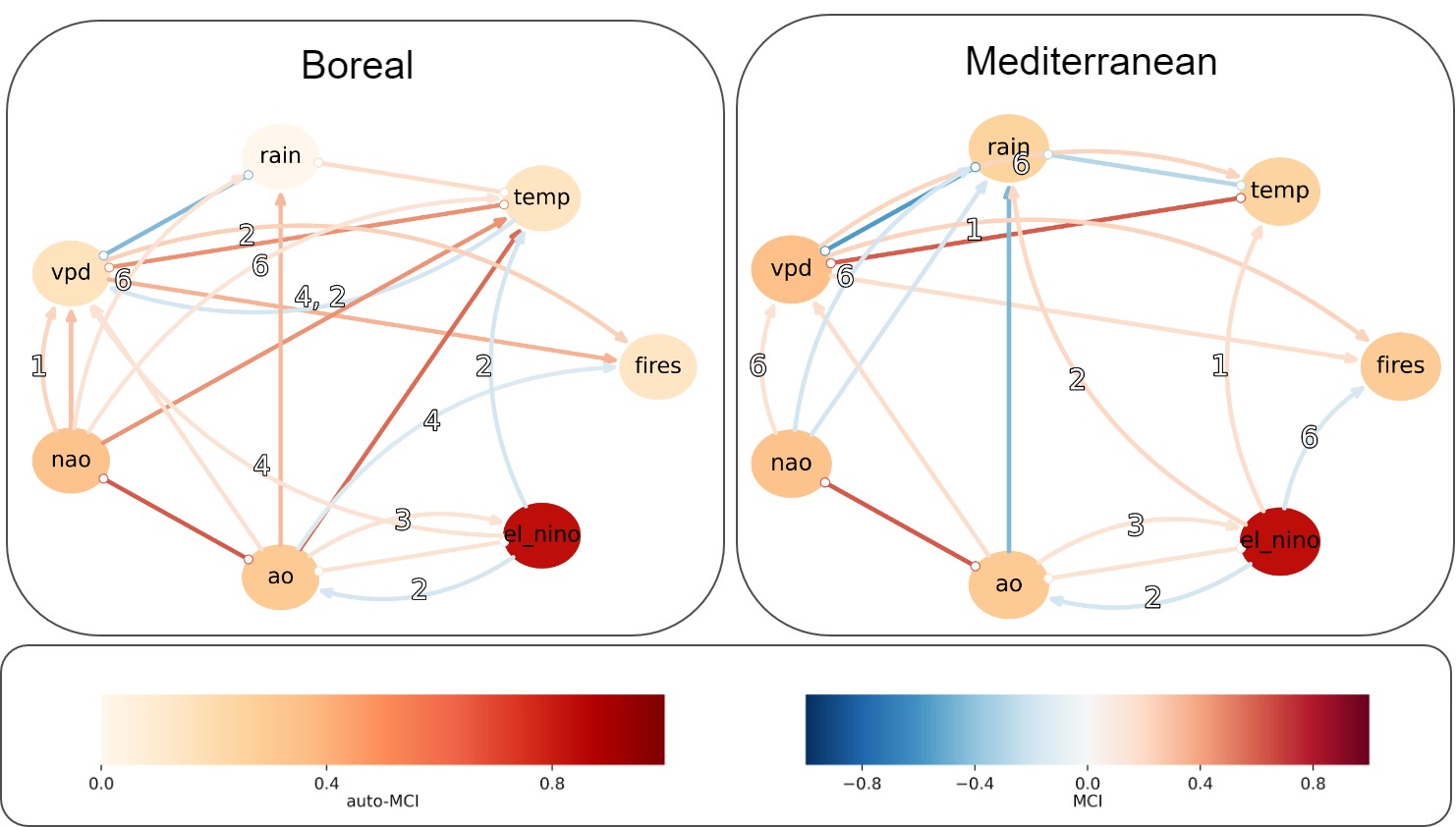} 
        \caption{Causal graphs over Europe's Boreal and Mediterranean biomes. The color of the lines (MCI) of the causal links shows the strength of the relationship (positive for reds and negative for blues). The numbers in the arrows show their lagged relationship in months, if missing then the relationship is contemporaneous (lag 0). The arrows show the directionality, if any. The color of the nodes (auto-MCI) shows the autocorrelation of each variable with the time series. }
        \label{fig:graphs}
\end{figure}

We compute causal graphs using PCMCI based on data from 2001 to 2019. The conditional independence test used is the one that accounts for partial correlations (ParCorr). The maximum time delay $\tau_{max}$ is six months. Significance level $\alpha$ at 0.05 is used to threshold the estimated matrix of p-values to get the graph. The derived causal graphs are shown in Fig.\ref{fig:graphs}.

\end{document}

%% file: math_commands.tex
\usepackage{amsmath,amsfonts,bm}

\def\eqref#1{equation~\ref{#1}}

\def\1{\bm{1}}

\DeclareMathAlphabet{\mathsfit}{\encodingdefault}{\sfdefault}{m}{sl}
\SetMathAlphabet{\mathsfit}{bold}{\encodingdefault}{\sfdefault}{bx}{n}













%% file: iclr2024_conference.bbl
\begin{thebibliography}{33}
\providecommand{\natexlab}[1]{#1}
\providecommand{\url}[1]{\texttt{#1}}
\expandafter\ifx\csname urlstyle\endcsname\relax
  \providecommand{\doi}[1]{doi: #1}\else
  \providecommand{\doi}{doi: \begingroup \urlstyle{rm}\Url}\fi

\bibitem[Batllori et~al.(2013)Batllori, Parisien, Krawchuk, and Moritz]{batllori_climate_2013}
Enric Batllori, Marc-André Parisien, Meg~A. Krawchuk, and Max~A. Moritz.
\newblock Climate change-induced shifts in fire for {Mediterranean} ecosystems.
\newblock \emph{Global Ecology and Biogeography}, 22\penalty0 (10):\penalty0 1118--1129, 2013.
\newblock ISSN 1466-8238.
\newblock \doi{10.1111/geb.12065}.
\newblock URL \url{https://onlinelibrary.wiley.com/doi/abs/10.1111/geb.12065}.
\newblock \_eprint: https://onlinelibrary.wiley.com/doi/pdf/10.1111/geb.12065.

\bibitem[Br\"{o}nnimann(2007)]{Brnnimann2007}
Stefan Br\"{o}nnimann.
\newblock Impact of el niño–southern oscillation on european climate.
\newblock \emph{Reviews of Geophysics}, 45\penalty0 (3), August 2007.
\newblock ISSN 1944-9208.
\newblock \doi{10.1029/2006rg000199}.
\newblock URL \url{http://dx.doi.org/10.1029/2006RG000199}.

\bibitem[Camps-Valls et~al.(2021)Camps-Valls, Tuia, Zhu, and Reichstein]{CampsValls21wiley}
Gustau Camps-Valls, Devis Tuia, XiaoXiang Zhu, and Markus Reichstein.
\newblock \emph{Deep learning for the Earth Sciences: A comprehensive approach to remote sensing, climate science and geosciences}.
\newblock Wiley \& Sons, 2021.
\newblock ISBN 9781119646143.
\newblock URL \url{https://github.com/DL4ES}.

\bibitem[Camps-Valls et~al.(2023)Camps-Valls, Gerhardus, Ninad, Varando, Martius, Balaguer-Ballester, Vinuesa, Diaz, Zanna, and Runge]{camps2023discovering}
Gustau Camps-Valls, Andreas Gerhardus, Urmi Ninad, Gherardo Varando, Georg Martius, Emili Balaguer-Ballester, Ricardo Vinuesa, Emiliano Diaz, Laure Zanna, and Jakob Runge.
\newblock Discovering causal relations and equations from data.
\newblock \emph{Physics Reports}, 1044:\penalty0 1--68, 2023.
\newblock ISSN 0370-1573.
\newblock \doi{https://doi.org/10.1016/j.physrep.2023.10.005}.

\bibitem[Cho et~al.(2014)Cho, Van~Merri{\"e}nboer, Gulcehre, Bahdanau, Bougares, Schwenk, and Bengio]{cho2014learning}
Kyunghyun Cho, Bart Van~Merri{\"e}nboer, Caglar Gulcehre, Dzmitry Bahdanau, Fethi Bougares, Holger Schwenk, and Yoshua Bengio.
\newblock Learning phrase representations using rnn encoder-decoder for statistical machine translation.
\newblock \emph{arXiv preprint arXiv:1406.1078}, 2014.

\bibitem[D{\'\i}az et~al.(2022)D{\'\i}az, Adsuara, Mart{\'\i}nez, Piles, and Camps-Valls]{diaz2022inferring}
Emiliano D{\'\i}az, Jose~E Adsuara, {\'A}lvaro~Moreno Mart{\'\i}nez, Mar{\'\i}a Piles, and Gustau Camps-Valls.
\newblock Inferring causal relations from observational long-term carbon and water fluxes records.
\newblock \emph{Scientific Reports}, 12\penalty0 (1):\penalty0 1610, 2022.

\bibitem[Gao et~al.(2023)Gao, Yang, Chen, Sugihara, Li, Stein, Kwan, and Wang]{gao2023causal}
Bingbo Gao, Jianyu Yang, Ziyue Chen, George Sugihara, Manchun Li, Alfred Stein, Mei-Po Kwan, and Jinfeng Wang.
\newblock Causal inference from cross-sectional earth system data with geographical convergent cross mapping.
\newblock \emph{nature communications}, 14\penalty0 (1):\penalty0 5875, 2023.

\bibitem[Hochreiter \& Schmidhuber(1997)Hochreiter and Schmidhuber]{memory2010long}
Sepp Hochreiter and Jürgen Schmidhuber.
\newblock Long short-term memory.
\newblock \emph{Neural computation}, 9:\penalty0 1735--80, 12 1997.
\newblock \doi{10.1162/neco.1997.9.8.1735}.

\bibitem[Iglesias-Suarez et~al.(2023)Iglesias-Suarez, Gentine, Solino-Fernandez, Beucler, Pritchard, Runge, and Eyring]{iglesias2023causally}
Fernando Iglesias-Suarez, Pierre Gentine, Breixo Solino-Fernandez, Tom Beucler, Michael Pritchard, Jakob Runge, and Veronika Eyring.
\newblock Causally-informed deep learning to improve climate models and projections.
\newblock \emph{arXiv preprint arXiv:2304.12952}, 2023.

\bibitem[Karasante et~al.(2023)Karasante, Alonso, Prapas, Ahuja, Carvalhais, and Papoutsis]{karasante2023seasfire}
Ilektra Karasante, Lazaro Alonso, Ioannis Prapas, Akanksha Ahuja, Nuno Carvalhais, and Ioannis Papoutsis.
\newblock Seasfire as a multivariate earth system datacube for wildfire dynamics.
\newblock \emph{arXiv preprint arXiv:2312.07199}, 2023.

\bibitem[Kondylatos et~al.(2022)Kondylatos, Prapas, Ronco, Papoutsis, Camps-Valls, Piles, Fern{\'a}ndez-Torres, and Carvalhais]{kondylatos2022wildfire}
Spyros Kondylatos, Ioannis Prapas, Michele Ronco, Ioannis Papoutsis, Gustau Camps-Valls, Mar{\'\i}a Piles, Miguel-{\'A}ngel Fern{\'a}ndez-Torres, and Nuno Carvalhais.
\newblock Wildfire danger prediction and understanding with deep learning.
\newblock \emph{Geophysical Research Letters}, 49\penalty0 (17):\penalty0 e2022GL099368, 2022.

\bibitem[Krich et~al.(2020)Krich, Runge, Miralles, Migliavacca, Perez-Priego, El-Madany, Carrara, and Mahecha]{Krich2020}
Christopher Krich, Jakob Runge, Diego~G. Miralles, Mirco Migliavacca, Oscar Perez-Priego, Tarek El-Madany, Arnaud Carrara, and Miguel~D. Mahecha.
\newblock Estimating causal networks in biosphere–atmosphere interaction with the pcmci approach.
\newblock \emph{Biogeosciences}, 17\penalty0 (4):\penalty0 1033–1061, February 2020.
\newblock ISSN 1726-4189.
\newblock \doi{10.5194/bg-17-1033-2020}.
\newblock URL \url{http://dx.doi.org/10.5194/bg-17-1033-2020}.

\bibitem[Li et~al.(2023)Li, Zhu, Riley, Zhao, Xu, Yuan, Chen, Wu, Gui, Gong, et~al.]{li2023attentionfire_v1}
Fa~Li, Qing Zhu, William~J Riley, Lei Zhao, Li~Xu, Kunxiaojia Yuan, Min Chen, Huayi Wu, Zhipeng Gui, Jianya Gong, et~al.
\newblock Attentionfire\_v1. 0: interpretable machine learning fire model for burned-area predictions over tropics.
\newblock \emph{Geoscientific Model Development}, 16\penalty0 (3):\penalty0 869--884, 2023.

\bibitem[Lundberg \& Lee(2017)Lundberg and Lee]{lundberg2017unified}
Scott~M Lundberg and Su-In Lee.
\newblock A unified approach to interpreting model predictions.
\newblock \emph{Advances in neural information processing systems}, 30, 2017.

\bibitem[M\"{u}ller-Plath et~al.(2022)M\"{u}ller-Plath, L\"{u}decke, and L\"{u}ning]{MllerPlath2022}
Gisela M\"{u}ller-Plath, Horst-Joachim L\"{u}decke, and Sebastian L\"{u}ning.
\newblock Long-distance air pressure differences correlate with european rain.
\newblock \emph{Scientific Reports}, 12\penalty0 (1), June 2022.
\newblock ISSN 2045-2322.
\newblock \doi{10.1038/s41598-022-14028-w}.
\newblock URL \url{http://dx.doi.org/10.1038/s41598-022-14028-w}.

\bibitem[Pearl(2009)]{Pearl2009}
Judea Pearl.
\newblock \emph{Causality: Models, Reasoning, and Inference}.
\newblock Cambridge University Press, September 2009.
\newblock ISBN 9780521749190.
\newblock \doi{10.1017/cbo9780511803161}.
\newblock URL \url{http://dx.doi.org/10.1017/CBO9780511803161}.

\bibitem[P{\'e}rez-Suay \& Camps-Valls(2018)P{\'e}rez-Suay and Camps-Valls]{perez2018causal}
Adri{\'a}n P{\'e}rez-Suay and Gustau Camps-Valls.
\newblock Causal inference in geoscience and remote sensing from observational data.
\newblock \emph{IEEE Transactions on Geoscience and Remote Sensing}, 57\penalty0 (3):\penalty0 1502--1513, 2018.

\bibitem[Pettinari \& Chuvieco(2020)Pettinari and Chuvieco]{pettinari2020fire}
M~Lucrecia Pettinari and Emilio Chuvieco.
\newblock Fire danger observed from space.
\newblock \emph{Surveys in Geophysics}, 41\penalty0 (6):\penalty0 1437--1459, 2020.

\bibitem[Prapas et~al.(2021)Prapas, Kondylatos, Papoutsis, Camps-Valls, Ronco, Fern{\'a}ndez-Torres, Guillem, and Carvalhais]{prapas2021deep}
Ioannis Prapas, Spyros Kondylatos, Ioannis Papoutsis, Gustau Camps-Valls, Michele Ronco, Miguel-{\'A}ngel Fern{\'a}ndez-Torres, Maria~Piles Guillem, and Nuno Carvalhais.
\newblock Deep learning methods for daily wildfire danger forecasting.
\newblock \emph{arXiv preprint arXiv:2111.02736}, 2021.

\bibitem[Prapas et~al.(2023)Prapas, Bountos, Kondylatos, Michail, Camps-Valls, and Papoutsis]{prapas2023televit}
Ioannis Prapas, Nikolaos-Ioannis Bountos, Spyros Kondylatos, Dimitrios Michail, Gustau Camps-Valls, and Ioannis Papoutsis.
\newblock Televit: Teleconnection-driven transformers improve subseasonal to seasonal wildfire forecasting.
\newblock In \emph{Proceedings of the IEEE/CVF International Conference on Computer Vision}, pp.\  3754--3759, 2023.

\bibitem[Reichstein et~al.(2013)Reichstein, Bahn, Ciais, Frank, Mahecha, Seneviratne, Zscheischler, Beer, Buchmann, Frank, et~al.]{reichstein2013climate}
Markus Reichstein, Michael Bahn, Philippe Ciais, Dorothea Frank, Miguel~D Mahecha, Sonia~I Seneviratne, Jakob Zscheischler, Christian Beer, Nina Buchmann, David~C Frank, et~al.
\newblock Climate extremes and the carbon cycle.
\newblock \emph{Nature}, 500\penalty0 (7462):\penalty0 287--295, 2013.

\bibitem[Reichstein et~al.(2019)Reichstein, Camps-Valls, Stevens, Jung, Denzler, Carvalhais, and Prabhat]{reichstein2019deep}
Markus Reichstein, Gustau Camps-Valls, Bjorn Stevens, Martin Jung, Joachim Denzler, Nuno Carvalhais, and fnm Prabhat.
\newblock Deep learning and process understanding for data-driven earth system science.
\newblock \emph{Nature}, 566\penalty0 (7743):\penalty0 195--204, 2019.

\bibitem[Runge et~al.(2019{\natexlab{a}})Runge, Bathiany, Bollt, Camps-Valls, Coumou, Deyle, Glymour, Kretschmer, Mahecha, Mu{\~n}oz-Mar{\'\i}, et~al.]{runge2019inferring}
Jakob Runge, Sebastian Bathiany, Erik Bollt, Gustau Camps-Valls, Dim Coumou, Ethan Deyle, Clark Glymour, Marlene Kretschmer, Miguel~D Mahecha, Jordi Mu{\~n}oz-Mar{\'\i}, et~al.
\newblock Inferring causation from time series in earth system sciences.
\newblock \emph{Nature communications}, 10\penalty0 (1):\penalty0 2553, 2019{\natexlab{a}}.

\bibitem[Runge et~al.(2019{\natexlab{b}})Runge, Nowack, Kretschmer, Flaxman, and Sejdinovic]{runge2019detecting}
Jakob Runge, Peer Nowack, Marlene Kretschmer, Seth Flaxman, and Dino Sejdinovic.
\newblock Detecting and quantifying causal associations in large nonlinear time series datasets.
\newblock \emph{Science advances}, 5\penalty0 (11):\penalty0 eaau4996, 2019{\natexlab{b}}.

\bibitem[Runge et~al.(2023)Runge, Gerhardus, Varando, Eyring, and Camps-Valls]{runge2023causal}
Jakob Runge, Andreas Gerhardus, Gherardo Varando, Veronika Eyring, and Gustau Camps-Valls.
\newblock Causal inference for time series.
\newblock \emph{Nature Reviews Earth \& Environment}, 4\penalty0 (7):\penalty0 487--505, 2023.

\bibitem[Saito \& Rehmsmeier(2015)Saito and Rehmsmeier]{saito2015precision}
Takaya Saito and Marc Rehmsmeier.
\newblock The precision-recall plot is more informative than the roc plot when evaluating binary classifiers on imbalanced datasets.
\newblock \emph{PloS one}, 10\penalty0 (3):\penalty0 e0118432, 2015.

\bibitem[Sch{\"o}lkopf et~al.(2021)Sch{\"o}lkopf, Locatello, Bauer, Ke, Kalchbrenner, Goyal, and Bengio]{scholkopf2021toward}
Bernhard Sch{\"o}lkopf, Francesco Locatello, Stefan Bauer, Nan~Rosemary Ke, Nal Kalchbrenner, Anirudh Goyal, and Yoshua Bengio.
\newblock Toward causal representation learning.
\newblock \emph{Proceedings of the IEEE}, 109\penalty0 (5):\penalty0 612--634, 2021.

\bibitem[Spirtes et~al.(2001)Spirtes, Glymour, and Scheines]{spirtes2001causation}
Peter Spirtes, Clark Glymour, and Richard Scheines.
\newblock \emph{Causation, Prediction, and Search}.
\newblock MIT Press, Cambridge, 2001.

\bibitem[Sugihara et~al.(2012)Sugihara, May, Ye, Hsieh, Deyle, Fogarty, and Munch]{sugihara2012detecting}
George Sugihara, Robert May, Hao Ye, Chih-hao Hsieh, Ethan Deyle, Michael Fogarty, and Stephan Munch.
\newblock Detecting causality in complex ecosystems.
\newblock \emph{science}, 338\penalty0 (6106):\penalty0 496--500, 2012.

\bibitem[Tsonis et~al.(2018)Tsonis, Deyle, Ye, and Sugihara]{tsonis2018convergent}
Anastasios~A Tsonis, Ethan~R Deyle, Hao Ye, and George Sugihara.
\newblock Convergent cross mapping: theory and an example.
\newblock \emph{Advances in nonlinear geosciences}, pp.\  587--600, 2018.

\bibitem[Varando et~al.(2021)Varando, Fern{\'a}ndez-Torres, and Camps-Valls]{varando2021learning}
Gherardo Varando, Miguel-Angel Fern{\'a}ndez-Torres, and Gustau Camps-Valls.
\newblock Learning granger causal feature representations.
\newblock In \emph{ICML 2021 Workshop on Tackling Climate Change with Machine Learning}, 2021.

\bibitem[Wang et~al.(2018)Wang, Yang, Chen, De~Maeyer, Li, and Duan]{wang2018detecting}
Yunqian Wang, Jing Yang, Yaning Chen, Philippe De~Maeyer, Zhi Li, and Weili Duan.
\newblock Detecting the causal effect of soil moisture on precipitation using convergent cross mapping.
\newblock \emph{Scientific reports}, 8\penalty0 (1):\penalty0 12171, 2018.

\bibitem[Zhang \& Hyvarinen(2012)Zhang and Hyvarinen]{zhang2012identifiability}
Kun Zhang and Aapo Hyvarinen.
\newblock On the identifiability of the post-nonlinear causal model.
\newblock \emph{arXiv preprint arXiv:1205.2599}, 2012.

\end{thebibliography}
